\documentclass[letterpaper]{article} 
\usepackage[draft]{aaai2026}    
\usepackage{times}                   
\usepackage{helvet}                  
\usepackage{courier}                 
\usepackage[hyphens]{url}            
\usepackage{graphicx}                
\usepackage{natbib}                  
\usepackage{caption}                 
\usepackage{iftex}
\ifXeTeX
  \input{t1enc.def}%
  \fontencoding{T1}\selectfont
  \usepackage{newunicodechar}%
  \newunicodechar{—}{\textemdash}\newunicodechar{–}{\textendash}%
  \newunicodechar{“}{``}\newunicodechar{”}{''}%
  \newunicodechar{‘}{`}\newunicodechar{’}{'}%
\fi

\usepackage{algorithm}
\usepackage{algorithmic}
\usepackage{amsmath}
\usepackage{amssymb}
\usepackage{amsthm}
\usepackage{booktabs}
\usepackage{multirow}
\usepackage{array}     
\usepackage{tikz}      
\usetikzlibrary{arrows.meta,positioning,fit,backgrounds,calc}
\usepackage{xcolor}

\newtheorem{proposition}{Proposition}

\newcommand{\method}{CRAFT}
\newcommand{\cti}{\widehat{\mathrm{CTI}}}
\newcommand{\sg}{\mathrm{sg}}
\newcommand{\rhog}{\rho_{\mathrm{gate}}}
\newcommand{\rhoema}{\rho_{\mathrm{gate}}^{\mathrm{EMA}}}
\newcommand{\Adv}{A}
\newcommand{\dt}{\Delta_t}
\newcommand{\ind}{\mathbf{1}}

\title{\method: Counterfactual Credit Assignment from Free Sibling\\
       Rollouts for Self-Distilled Agentic Reinforcement Learning}

\author{
    Zibin Meng,
    Kani Chen\textsuperscript{*}
}
\affiliations{
    The Hong Kong University of Science and Technology\\
    zmengal@connect.ust.hk, makchen@ust.hk\\
    \textsuperscript{*}Corresponding author.
}

\begin{document}
\maketitle

\begin{abstract}
Self-distilled agentic reinforcement learning augments
trajectory-level reward with a token-level distillation loss, using as
its teacher the same policy conditioned on privileged context. The
prevailing recipe gates this loss by a single scalar, the
teacher--student log-probability gap~$\dt$. We observe that this signal
is doubly limited: it is \emph{retrospective}, scoring only the
realised rollout and never the counterfactual ones, and it is
\emph{sign-blind}, never signalling when a teacher-preferred action
would have \emph{harmed} the trajectory.
We introduce~\method, a three-pillar credit-assignment scheme that
addresses both limitations. \textbf{Pillar~1}, Counterfactual Token
Importance, reuses the $G-1$ sibling rollouts that GRPO already samples
and importance-weights them by~$\dt$ to form a self-normalised estimate
of the group-level counterfactual change in advantage from up-weighting
teacher-preferred actions at each step; this yields a signed per-token
credit at essentially no extra compute. \textbf{Pillar~2} is an
asymmetric controller that raises the distillation weight as it lowers
the reference-KL weight along an exponential moving average of gate
activity, and conversely. \textbf{Pillar~3} polarises the KL penalty
token by token, switching between a mode-seeking and a mode-covering
update according to the sign of the credit.
Each pillar is governed by an independent switch that, when disabled,
renders the loss and gradient byte-identical to the baseline in
IEEE-754 arithmetic, so any measured gain is attributable to
algorithmic change rather than implementation drift. We establish the
estimator's consistency and a variance bound, give structural and
bit-exact reproducibility guarantees, and evaluate~\method{} across
three agentic environments, four model scales, and five end-to-end
methods, plus two tabulated prior-work baselines. Among these is
\emph{Adaptive-CRINGE}, a comparator that shares Pillar~2
with~\method{} and thereby isolates the counterfactual contribution.
\end{abstract}


\section{Introduction}
\label{sec:intro}

Agentic reinforcement learning trains a language-model policy
$\pi_\theta$ to interact with a tool-using or web-acting environment
over many turns~\cite{shridhar2021alfworld,yao2022webshop,jin2024searchr1},
building on a recent line of LLM-as-agent methods that compose
reasoning, tool use, and self-critique~\cite{yao2022react,
schick2023toolformer,shinn2023reflexion}.
Two ingredients dominate the field. (1)~The policy is updated with a
group-relative advantage estimator
(GRPO~\cite{shao2024grpo}), in which $G$ parallel rollouts of the same
prompt produce a normalised, sequence-level advantage~$\Adv^{(i)}$.
(2)~On top of that surrogate, an auxiliary self-distillation loss
treats the same policy, conditioned on additional privileged context
$s^+$ (a skill annotation, a verifier hint, or a retrieved trajectory
trace), as a~\emph{teacher} $\pi_T$. The teacher-student log-prob gap
$\dt = \log\pi_T(y_t \mid s_t^+) - \log\pi_\theta(y_t \mid s_t)$ is
then fed to a single gated distillation term, of the form
$\sigma(\beta\,\dt) \cdot (\log\pi_T - \log\pi_\theta)$, which the
recent line of work on \emph{Self-Distilled Agentic
Reinforcement Learning} (SDAR~\cite{lu2026sdar}) has popularised.

\paragraph{What is missing.}
The single-gate recipe quietly conflates two qualitatively different
signals into one scalar:

\begin{itemize}
\item \textbf{Retrospective vs.\ counterfactual credit.}
$\dt$ records what the teacher and student said on the
\emph{realised} trajectory. Two tokens with the same $\dt$ but
opposite impact on the trajectory's return are weighted identically.
The signal contains \emph{no} forward-looking information about what
would have happened had the agent followed the teacher's preference
at step~$t$ — even though GRPO has already paid the compute to
sample $G$ counterfactual rollouts.
\item \textbf{Sign-blindness.}
The gate $\sigma(\beta\,\dt)$ is monotone in $\dt$.
Tokens where $\pi_T > \pi_\theta$ get up-weighted forward KL; tokens
where $\pi_T < \pi_\theta$ get a small but still~\emph{positive} gate.
The loss never tells the policy ``this teacher action would have been
\emph{worse}; do not follow it.''
\end{itemize}

These weaknesses motivate a richer per-token signal, but the design
space is treacherous. Pushing the policy away from the teacher whenever
the gap is negative recovers
unlikelihood~\cite{welleck2020unlikelihood} or
CRINGE~\cite{adolphs2022cringe}, both of which are well-known to over-
penalise and to ignore reward signal entirely. Reweighting by the
trajectory-level $\mathrm{sign}(\Adv^{(i)}) \cdot \dt$ recovers the
RLSD line of work~\cite{yang2026rlsd},
which is \emph{retrospective} (it conditions on the realised
trajectory, not on a counterfactual one). And blindly stacking an
adaptive KL controller~\cite{schulman2017ppo,ouyang2022instructgpt} on
top of SDAR adjusts a single coefficient without changing the per-token
gating discipline.

\paragraph{Our position.}
We argue that the right per-token credit signal in self-distilled
agentic RL is the \emph{counterfactual change in sequence advantage}
that the policy would have realised if, at step~$t$, it had sampled
from the teacher. We further argue that GRPO already provides the
sample budget needed to estimate that counterfactual: the $G-1$
\emph{sibling rollouts} in the same group, importance-weighted by the
teacher-student gap at step~$t$, form an essentially free
importance-sampling estimator of that counterfactual at the group
level (made precise in Section~\ref{sec:pillar1}). Building on this
estimator we propose \textbf{\method}, a three-pillar
credit-assignment scheme detailed in Section~\ref{sec:method}.

\paragraph{Contributions.}
Each contribution is gated by an independent master switch, so
ablations factor cleanly.
\textbf{(1)~Pillar~1 (CTI):} a per-token signed credit
$c_t\in(-1,1)$ from importance-weighted sibling rollouts; the
underlying estimator is consistent (Prop.~\ref{prop:p1}), has an
$O(1/G)$ variance bound (Prop.~\ref{prop:p2}), and reduces to the
single-gate baseline in the limit (P3) --- at the marginal cost of one
group-wise softmax.
\textbf{(2)~Pillar~2:} a bivariate controller driving $\lambda(t)$
(CTI) and $\mu(t)$ (reference-KL) in \emph{opposite directions} along
the gate-active EMA, with a fixed point (P4).
\textbf{(3)~Pillar~3:} a per-token KL penalty whose direction is set
by $\mathrm{sign}(c_t)$ (mode-seeking vs.\ mode-covering),
structurally consistent with the credit (P5) and yielding a valid
policy-gradient surrogate (P6).
\textbf{(4)~Bit-exact reproducibility:} with all switches off, loss
and gradient are byte-identical (IEEE-754) to the baseline (P7), so
gains are attributable to algorithmic change, not drift.
\textbf{(5)~A campaign} over three agentic environments, four model
sizes, and four baselines --- including the hard \emph{Adaptive-CRINGE}
comparator that shares Pillar~2 with~\method --- plus ablations and
OOD evaluation (Section~\ref{sec:exp}).


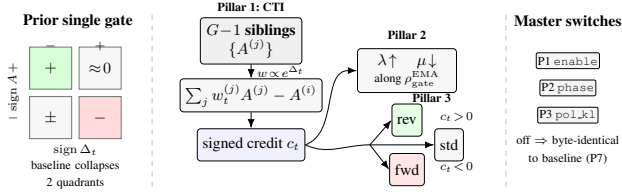
\begin{figure}[t]
\centering
\resizebox{\linewidth}{!}{%
\begin{tikzpicture}[
  font=\footnotesize, >=Latex, line width=0.5pt,
  qcell/.style={draw=black!55, minimum size=8.5mm, inner sep=1pt, align=center},
  blk/.style={draw, rounded corners=2pt, align=center, inner sep=3pt,
              minimum height=6.5mm, fill=black!3},
  pill/.style={draw, rounded corners=2pt, align=center, inner sep=3.5pt,
               font=\footnotesize\bfseries, fill=black!6, minimum height=6mm},
  sw/.style={draw, rounded corners=1pt, minimum width=6mm, minimum height=3.2mm,
             inner sep=1pt, font=\scriptsize}
]
\node[font=\footnotesize\bfseries] at (1.05,2.55) {Prior single gate};
\node[qcell, fill=green!16] (q11) at (0.5,1.6)  {$+$};
\node[qcell, fill=black!3]  (q01) at (1.5,1.6)  {$\approx\!0$};
\node[qcell, fill=black!3]  (q10) at (0.5,0.6)  {$\pm$};
\node[qcell, fill=red!14]   (q00) at (1.5,0.6)  {$-$};
\node[rotate=90, font=\scriptsize] at (-0.25,1.1) {$\mathrm{sign}\,\Adv$};
\node[font=\scriptsize] at (1.0,-0.05) {$\mathrm{sign}\,\dt$};
\node[font=\scriptsize] at (0.5,2.05) {$-$}; \node[font=\scriptsize] at (1.5,2.05) {$+$};
\node[font=\scriptsize, anchor=east] at (-0.02,1.6) {$+$};
\node[font=\scriptsize, anchor=east] at (-0.02,0.6) {$-$};
\node[font=\scriptsize, text width=2.3cm, align=center] at (1.0,-0.55)
  {baseline collapses 2 quadrants};

\node[pill] (sib) at (4.55,2.15) {$G\!-\!1$ siblings\\$\{\Adv^{(j)}\}$};
\node[blk]  (cti) at (4.55,1.05) {$\sum_{j}w^{(j)}_t\Adv^{(j)}-\Adv^{(i)}$};
\node[blk, fill=blue!6] (ct) at (4.55,0.05) {signed credit $c_t$};
\draw[->] (sib) -- (cti) node[midway,right,font=\scriptsize] {$w\!\propto\! e^{\dt}$};
\draw[->] (cti) -- (ct);
\node[font=\scriptsize\bfseries, anchor=south] at (4.55,2.62) {Pillar 1: CTI};

\node[blk, minimum width=20mm] (p2) at (7.7,1.55)
  {$\lambda\!\uparrow\quad\mu\!\downarrow$\\\scriptsize along $\rhoema$};
\node[font=\scriptsize\bfseries, anchor=south] at (7.7,2.05) {Pillar 2};
\draw[->] (ct.east) to[out=20,in=180] (p2.west);

\node[blk, fill=green!10] (rev) at (7.7,0.55) {rev};
\node[blk] (std) at (8.55,0.05) {std};
\node[blk, fill=red!10] (fwd) at (7.7,-0.45) {fwd};
\node[font=\scriptsize\bfseries] at (8.2,1.0) {Pillar 3};
\draw[->] (ct.east) to[out=-25,in=180] (6.95,0.05) -- (rev.west);
\draw[->] (6.95,0.05) -- (fwd.west);
\draw[->] (6.95,0.05) -- (std.west);
\node[font=\scriptsize] at (8.7,0.62) {$c_t\!>\!0$};
\node[font=\scriptsize] at (8.7,-0.4) {$c_t\!<\!0$};

\node[font=\footnotesize\bfseries] at (10.95,2.55) {Master switches};
\node[sw, fill=black!4] (s1) at (10.95,1.75) {P1 \texttt{enable}};
\node[sw, fill=black!4] (s2) at (10.95,1.20) {P2 \texttt{phase}};
\node[sw, fill=black!4] (s3) at (10.95,0.65) {P3 \texttt{pol\_kl}};
\node[font=\scriptsize, text width=2.6cm, align=center] at (10.95,-0.05)
  {off $\Rightarrow$ byte-identical to baseline (P7)};

\draw[black!25, dashed] (2.55,-0.8) -- (2.55,2.7);
\draw[black!25, dashed] (9.7,-0.8) -- (9.7,2.7);
\end{tikzpicture}}
\caption{
\textbf{The three pillars of~\method{} at a glance.}
\textit{Left:} on the
$(\mathrm{sign}\,\Adv^{(i)}, \mathrm{sign}\,\dt)$ plane, the prior
single-gate baseline collapses two of the four quadrants by always
distilling toward the teacher (Table~\ref{tab:quadrants}).
\textit{Centre:} Pillar~1 estimates a counterfactual credit
$\cti^{(i)}_t = \sum_{j\ne i} w_t^{(j)}\Adv^{(j)} - \Adv^{(i)}$
from the $G-1$ sibling rollouts of the same GRPO group, squashed
into $c_t\in(-1,1)$ by centred asymmetric sigmoids; Pillar~2 drives
$(\lambda,\mu)$ in opposite directions along the gate-activity
EMA; Pillar~3 routes the KL penalty into a rev / fwd / standard
branch according to $\mathrm{sign}(c_t)$.
\textit{Right:} each pillar has an independent master switch that,
when off, restores byte-identical behaviour with the prior
baseline (P7).
}
\label{fig:hero}
\end{figure}

\section{Preliminaries}
\label{sec:prelim}

\paragraph{GRPO and group-based advantage.}
For each prompt we sample $G$ parallel rollouts under the
\emph{behaviour} policy~$\pi_b$ (in on-policy GRPO,
$\pi_b = \pi_\theta$). Each rollout $i$ receives a scalar
sequence-level reward $r^{(i)}$ from a verifier; the per-trajectory
GRPO advantage~$\Adv^{(i)} \in \mathbb{R}$ is the
group-normalised reward, broadcast over all valid tokens in the
trajectory~\cite{shao2024grpo}; GRPO is also the policy-update
primitive in subsequent large-scale reasoning
models~\cite{deepseek2024v3}. The policy update follows the
PPO-clipped surrogate~\cite{schulman2017ppo}:
\begin{equation}
\begin{aligned}
L_{\mathrm{PG}}(\theta) = \mathbb{E}\Big[\min\Big(&r_t(\theta)\,\Adv^{(i)},\\
&\mathrm{clip}(r_t(\theta), 1\!-\!\epsilon, 1\!+\!\epsilon)\,\Adv^{(i)}\Big)\Big],
\end{aligned}
\end{equation}
where $r_t(\theta) = \pi_\theta(y_t \mid s_t) / \pi_{\theta_{\mathrm{old}}}(y_t \mid s_t)$.

\paragraph{Self-distilled teacher.}
A second forward pass evaluates the same policy on the
\emph{skill-augmented context}~$s_t^+$ — typically the prompt plus an
oracle skill annotation, a verifier hint, or a retrieved successful
trajectory — and yields the \emph{teacher} log-prob
$\log\pi_T(y_t \mid s_t^+)$. Because the teacher is just the student
under a different context, it adds a single forward pass (no extra
parameters, no extra optimiser state) and is gradient-free. The
teacher-student gap
\begin{equation}
\dt^{(i)} = \sg\!\left[\log\pi_T(y_t^{(i)} \mid s_t^{+,(i)}) -
                    \log\pi_\theta(y_t^{(i)} \mid s_t^{(i)})\right]
\end{equation}
is the substrate for the prior single-gate SDAR distillation loss:
\begin{equation}
L_{\mathrm{SDAR}} = \mathrm{Agg}\!\left[\,
   \underbrace{\sigma(\beta\,\dt)}_{\mathrm{gate}\;g_t}
   \cdot
   \underbrace{(\log\pi_T - \log\pi_\theta)}_{\mathrm{fwd-KL\ surrogate}}\,\right].
\label{eq:sdar}
\end{equation}

Throughout, $i,j\in\{1,\dots,G\}$ index siblings, $\sg[\cdot]$ is the
stop-gradient, $\ind[\cdot]$ the indicator, and $\pi_{\mathrm{ref}}$ a
frozen snapshot of $\pi_\theta$ serving as the KL anchor; a full
notation table is in Appendix~C.

\paragraph{Assumptions used in the propositions.}
\textbf{(A1)~Coverage:} for every $(s,y)$ with $\pi_T(y\mid s) > 0$,
$\pi_b(y\mid s) > 0$.
\textbf{(A2)~Lipschitz log-prob:}
$|\log\pi_\theta(y\mid s) - \log\pi_\theta(y\mid s')| \le L\,\lVert s-s'\rVert$.
\textbf{(A3)~Bounded advantages:} $|\Adv^{(i)}| \le R_{\max}$ a.s.
\textbf{(A4)~Group i.i.d.\@:} siblings in a group are independently
sampled from $\pi_b$ given the prompt.

\section{The \method{} Method}
\label{sec:method}

The three pillars sit on top of an unchanged $L_{\mathrm{PG}}$ and an
unchanged teacher pass. Concretely the composed loss is
\begin{equation}
\boxed{\;
\begin{aligned}
L_{\method}(\theta) ={}& L_{\mathrm{PG}}(\theta)
\;+\; \mu(t)\,L_{\mathrm{KL\text{-}pol}}(\theta)\\
&{}+\; \lambda(t)\,L_{\mathrm{CTI}}(\theta).
\end{aligned}
\;}
\label{eq:full-loss}
\end{equation}
The new per-token loss $L_{\mathrm{CTI}}$ (Section~\ref{sec:pillar1}),
the coefficients $\lambda(t),\mu(t)$ from the Pillar-2 controller
(Section~\ref{sec:pillar2}), and the polarised KL
$L_{\mathrm{KL\text{-}pol}}$ (Section~\ref{sec:pillar3}) can each be
disabled independently via \texttt{vera.\{enable,phase\_aware,%
polarized\_kl\}}, reducing to baseline behaviour (Sec.~\ref{sec:theory}).
The pillars are jointly necessary: Pillar~1 produces the signed
credit~$c_t$ --- the only quantity mixing both
$\mathrm{sign}\,\Adv^{(i)}$ and $\mathrm{sign}\,\dt$ --- whose
\emph{magnitude} ($\rhog$) Pillar~2 uses to schedule the weights and
whose \emph{sign} Pillar~3 uses to route the KL penalty.

\subsection{Pillar 1: Counterfactual Token Importance (CTI)}
\label{sec:pillar1}

\paragraph{Idealised per-trajectory target.}
The quantity we would \emph{ideally} like at position $t$ of
trajectory $i$ is the \emph{per-trajectory counterfactual token
importance}
\begin{equation}
\mathrm{CTI}^{(i)}_t \;:=\; \mathbb{E}_{y'_t \sim \pi_T(\cdot\mid s_t^{+,(i)})}\!\!\left[
A\!\left(\tau^{(i)}_{<t},\,y'_t,\,\tau^{(i)}_{>t}\right)\right]
- \Adv^{(i)},
\label{eq:cti-pop}
\end{equation}
the change in expected sequence-level advantage if at step~$t$
trajectory~$i$ had drawn its next token from the teacher and re-rolled
its continuation. This is not directly estimable from realised
rollouts --- it would require re-rolling $i$ from step~$t$ under fresh
teacher draws, a cost we do not pay. We instead estimate a
\emph{group-level} counterfactual that GRPO's existing siblings
support exactly, and connect it to~\eqref{eq:cti-pop} by an explicit
assumption.

\paragraph{Group-level estimand and sibling-pool estimator.}
Given the prompt, the $G-1$ siblings $j\ne i$ are i.i.d.\ draws of the
step-$t$ triple $(s_t^{(j)},y_t^{(j)},\Adv^{(j)})$ from the behaviour
law $\mu_t$. Tilting that law toward tokens the teacher prefers
(relative to the student) defines the \emph{teacher-tilted group
advantage} and the group-level counterfactual importance
\begin{equation}
\overline{\mathrm{CTI}}^{(i)}_t \;:=\;
\frac{\mathbb{E}_{\mu_t}[\varrho\,\Adv]}{\mathbb{E}_{\mu_t}[\varrho]}
\;-\; \Adv^{(i)},
\qquad
\varrho := \frac{\pi_T(y_t\mid s_t^{+})}{\pi_\theta(y_t\mid s_t)},
\label{eq:cti-group}
\end{equation}
the group-mean advantage after up-weighting teacher-preferred
step-$t$ actions, minus the realised advantage of~$i$. We stress that
the expectations in~\eqref{eq:cti-group} are taken over the group's
\emph{joint} step-$t$ law $\mu_t$ on $(s_t,y_t,\Adv)$: because
siblings reach step $t$ through different prefixes, $\mu_t$ marginalises
over heterogeneous states, so $\overline{\mathrm{CTI}}^{(i)}_t$ is a
\emph{group-level, state-marginal} tilt rather than a fixed-state
intervention on trajectory~$i$. The identification with the
per-trajectory object~\eqref{eq:cti-pop} is exactly what
assumption~(E) below supplies. We estimate~\eqref{eq:cti-group}
by the self-normalised importance-weighted average over the siblings:
\begin{equation}
\begin{aligned}
\cti^{(i)}_t &\;=\; \sum_{j \ne i} w^{(j)}_t \, \Adv^{(j)} \;-\; \Adv^{(i)},\\
w^{(j)}_t &\;=\; \frac{\exp(\dt^{(j)}/\tau)}{\sum_{k \ne i}\exp(\dt^{(k)}/\tau)}.
\end{aligned}
\label{eq:cti-est}
\end{equation}
At the default $\tau=1$ the weight is the per-sibling teacher/student
ratio $w^{(j)}_t \propto \exp(\dt^{(j)}) = \varrho^{(j)}$ (with
$\mathbb{E}_{\mu_t}[\varrho\mid s_t]=1$), making~\eqref{eq:cti-est} the
textbook self-normalised importance-sampling (SNIS) estimator of the
target~\eqref{eq:cti-group}, consistent as $G\!\to\!\infty$
(Prop.~\ref{prop:p1}). The temperature traces a bias-variance
trade-off around this point ($\tau\!\to\!0$: winner-take-all, high
variance; $\tau\!\to\!\infty$: uniform, $\dt$-agnostic), so we fix
$\tau=1$ and sweep $\{0.5,2\}$ only as a robustness check
(Ablation~A5).

\paragraph{From group-level to per-trajectory.}
The group-level estimand~\eqref{eq:cti-group} coincides with the
per-trajectory target~\eqref{eq:cti-pop} under
\textbf{(E)~within-group exchangeability}: the spliced and a sibling's
realised advantage share the same conditional mean given $(s_t,y_t)$.
(E) holds \emph{exactly} at $t=0$ and degrades smoothly downstream.
Every proposition is stated for the group-level
estimand~\eqref{eq:cti-group} and so does \emph{not} rely on~(E); we
read $\cti^{(i)}_t$ as an \emph{exact} group-level and
\emph{approximate} per-trajectory credit, with Ablation~A6
($G_{\min}{=}999$) the empirical control. For groups below $G_{\min}$
we fall back to the single-rollout proxy
$\cti^{(i)}_t\!\leftarrow\!\Adv^{(i)}\dt^{(i)}$ (or $0$), a sign-aware
variant of SDAR gating --- the limiting case of P3.

\paragraph{Signed credit.}
The per-token signed credit is
\begin{align}
c_t \;&=\;
\bigl(2\sigma(\beta_+ \cti_t)-1\bigr)\,\ind[\cti_t > 0]\nonumber\\
&\quad\;-\;
\bigl(2\sigma(\beta_- |\cti_t|)-1\bigr)\,\ind[\cti_t < 0],
\label{eq:signed-credit}
\\
c_t \;&=\; \sg[c_t] \in (-1, 1).
\end{align}
The \emph{centred} sigmoids ($2\sigma(\cdot)-1$) make $c_t$ continuous
at $\cti_t=0$ with a near-zero neutral band (where Pillar~3 falls back
to standard KL), saturating to $\pm1$ as $|\cti_t|\to\infty$; the
asymmetric $\beta_+,\beta_-$ tune positive/negative reinforcement
independently. The credit is fully detached: gradients flow only
through $\log\pi_\theta$.

\paragraph{The CTI loss.}
With $c_t$ in hand, the Pillar~1 loss is the standard REINFORCE-style
policy-gradient form~\cite{williams1992reinforce,sutton1999pg}
treating $c_t$ as the credit:
\begin{equation}
L_{\mathrm{CTI}}(\theta) \;=\; -\,\mathrm{Agg}\!\left[\;
c_t \cdot \log\pi_\theta(y_t \mid s_t)\;\right].
\label{eq:cti-loss}
\end{equation}
The aggregator $\mathrm{Agg}[\cdot]$ matches the surrounding policy
loss (token-mean by default).

\paragraph{Marginal compute cost.}
The CTI estimator adds no model evaluation: it runs under
\texttt{torch.no\_grad()} as a group-wise softmax plus a weighted sum
over the $G-1$ sibling advantages ($O(B\,T\,G)$ pointwise tensor
operations), so its marginal cost over the prior single-gate baseline
is negligible.

\paragraph{Why CTI is not unlikelihood / CRINGE / RLSD.}
Unlikelihood~\cite{welleck2020unlikelihood} and
CRINGE~\cite{adolphs2022cringe} apply \emph{unconditional} token-level
repulsion and ignore reward; RLSD~\cite{yang2026rlsd} reweights by the
trajectory's own $\mathrm{sign}(\Adv^{(i)})\cdot\dt$. All share the
\emph{form} of a per-token weight on a log-probability term but differ
in what \emph{conditions} it: as Table~\ref{tab:signal-comparison}
shows, CTI is the only one that combines both signs
$(\mathrm{sign}\,\Adv^{(i)},\mathrm{sign}\,\dt)$ \emph{jointly} via a
counterfactual estimator over sibling rollouts.

\begin{table}[t]
\centering
\small
\setlength{\tabcolsep}{3pt}
\newcommand{\xm}{{\large{\textendash}}}
\begin{tabular}{@{}>{\raggedright\arraybackslash}p{0.46\columnwidth}cccc@{}}
\toprule
\textbf{Property} & \textbf{CRINGE} & \textbf{RLSD} & \textbf{SDAR} & \textbf{Ours} \\
\midrule
Uses reward signal $\Adv$           & \xm & \checkmark & \checkmark & \checkmark \\
Uses teacher gap $\dt$              & \checkmark & \checkmark & \checkmark & \checkmark \\
Counterfactual (forward-looking)    & \xm & \xm & \xm & \checkmark \\
Sign-aware (neg.\ branch fires)  & \checkmark & \checkmark & \xm & \checkmark \\
Bounded credit $|c_t|\le 1$         & \checkmark & \xm & \checkmark & \checkmark \\
Uses sibling pool / group structure & \xm & \xm & \xm & \checkmark \\
\bottomrule
\end{tabular}
\caption{Structural signal comparison of per-token credit-assignment
losses for self-distilled RL (\checkmark~= has the property,
\xm~= does not). The first two properties are necessary
but not sufficient (every reasonable per-token signal uses them) and
are read at the level of the full method: SDAR's per-token gate is
$\dt$-only while its policy-gradient term supplies~$\Adv$, and CRINGE's
$\dt$ entry denotes its token-level disagreement signal rather than an
explicit teacher gap. The last four properties together pick out the
\method{} design point.}
\label{tab:signal-comparison}
\end{table}

\paragraph{The 4-quadrant decision view.}
A token's contribution to the CTI estimator is determined jointly by
$\mathrm{sign}\,\Adv^{(i)}$ and $\mathrm{sign}\,\dt$, carving out the
four quadrants of Table~\ref{tab:quadrants} --- the natural partition
of the estimator, not a design choice. The two agreement quadrants
($\mathrm{sign}\,\Adv=\mathrm{sign}\,\dt$) push $c_t$ toward $\pm1$,
whereas the two disagreement quadrants either hold near zero or flip
sign according to the sibling pool. Collapsing them, e.g.\ by using
$|\cti|$ only, destroys the asymmetry that Pillar~3 exploits, while a
near-zero neutral band is exposed as the \texttt{vera.gate\_threshold}
knob (Ablation~A5).

\subsection{Pillar 2: Phase-Aware Adaptive Controller}
\label{sec:pillar2}

\paragraph{Motivation.}
Distillation moves through an \emph{active} phase (high $\rhog$, teacher
has unabsorbed signal), a \emph{settling} phase, and a \emph{saturated}
phase (low $\rhog$). A static CTI weight either over-distils when
saturated or under-distils when active, and a static KL coefficient
over-anchors early when the policy must move fast.

\paragraph{Control law.}
Let $\rhog(t)$ be the fraction of valid tokens in the current
mini-batch with $|c_t| > \theta_g$. The controller maintains an
exponential moving average (EMA)
\begin{equation}
\rhoema(t+1) \;=\; \alpha\,\rhoema(t) \;+\; (1-\alpha)\,\rhog(t),
\end{equation}
and produces, at step $t$,
\begin{align}
\lambda(t) &= \lambda_{\min} + (\lambda_{\max} - \lambda_{\min})\,\rhoema(t),
\label{eq:lam-law}\\
\mu(t) &= \mu_{\min} + (\mu_{\max} - \mu_{\min})\,\bigl(1 - \rhoema(t)\bigr).
\label{eq:mu-law}
\end{align}
The two coefficients move in \emph{opposite directions}; during early
training $\lambda$ rises (absorb teacher signal aggressively) while
$\mu$ falls (loosen the KL anchor); late in training the roles
invert. During the first $n_{\mathrm{warmup}}$ steps the
controller holds its initial mid-range state, preventing early-step
noise from setting a bad trajectory.

\paragraph{Differences from adaptive KL.}
Unlike adaptive-KL controllers~\cite{schulman2017ppo,ouyang2022instructgpt}
that adjust a single coefficient by measured KL distance, Pillar~2
moves \emph{two coupled} coefficients in \emph{opposite directions}
(a Pareto trade-off between distillation throughput $\lambda\rhog$ and
exploration budget $\mu(1{-}\rhog)$), driven by the
\emph{distillation-quality} signal $\rhog$ --- token-level
teacher--student disagreement, not distance from a fixed reference ---
with an EMA fixed point at $\rhog\to\bar\rho$ (P4).
The controller is a purely numeric state object updated once per step
from Pillar~1's $\rhog$ (Algorithm~\ref{alg:adaptive}, Appendix~B).

\subsection{Pillar 3: Direction-Polarised KL}
\label{sec:pillar3}

\paragraph{Motivation.}
The prior recipe applies one KL form
($k_3$~\cite{schulman2020kl}) to every token. The signed credit lets
us do better: where $c_t>0$ (following the teacher would have helped)
the policy should \emph{absorb} (mode-seeking); where $c_t<0$ (it would
have hurt) it should \emph{preserve coverage} (mode-covering); where
$|c_t|\approx0$ it falls back to standard KL. A single KL form cannot
express this asymmetry.

\paragraph{The three-region rule.}
With threshold $\theta_{KL} \ge 0$:
\begin{align}
L_{\mathrm{KL\text{-}pol}} = \mathrm{Agg}\Big[\;
& \ind[c_t > \theta_{KL}]\;\mathrm{rev}(t) \nonumber\\
+\;& \ind[c_t < -\theta_{KL}]\;\mathrm{fwd}(t) \nonumber\\
+\;& \ind[|c_t| \le \theta_{KL}]\;\mathrm{std}(t)\;\Big],
\label{eq:polkl}
\end{align}
where
\begin{align*}
\mathrm{rev}(t) &= \log\pi_\theta(y_t\mid s_t) - \log\pi_{\mathrm{ref}}(y_t\mid s_t),\\
\mathrm{fwd}(t) &= -\bigl(\log\pi_\theta(y_t\mid s_t) - \log\pi_{\mathrm{ref}}(y_t\mid s_t)\bigr),\\
\mathrm{std}(t) &= \mathrm{kl\_penalty}\!\left(\log\pi_\theta, \log\pi_{\mathrm{ref}}\right).
\end{align*}
Here $\mathrm{rev}(t)$ is the $k_1$ reverse-KL log-ratio (minimising it
pulls toward $\pi_{\mathrm{ref}}$, mode-seeking), $\mathrm{fwd}(t)$ its
sign-flip (pushes away, mode-covering), and $\mathrm{std}(t)$ the
standard $k_3$ penalty. The forward branch is \emph{not} an unbiased
estimator of $\mathrm{KL}(\pi_{\mathrm{ref}}\!\parallel\!\pi_\theta)$
--- the IS-weighted alternative has unbounded variance exactly where it
fires hardest --- so we adopt it as a policy-update rule inspired by
the forward-KL direction, not a divergence estimator (Appendix~B
compares three options).

\section{Theoretical Analysis}
\label{sec:theory}

Our analysis centres on one substantive result --- the CTI estimator
with its consistency and $O(1/G)$ variance bound
(Props.~\ref{prop:p1}--\ref{prop:p2}) --- supported by elementary
structural facts that make the design self-consistent. Full proofs are
in Appendix~B; where a result's strongest form exceeds what we prove
rigorously we state the weaker form the argument supports, and each
result is paired with a unit or end-to-end test.

\begin{proposition}[CTI estimator: consistency]
\label{prop:p1}
Fix the temperature at $\tau=1$, so that $w^{(j)}_t$
in~\eqref{eq:cti-est} is the self-normalised importance weight over
the sibling pool. Under (A1) (coverage) and (A4) (group i.i.d.),
$\cti^{(i)}_t$ is a self-normalised importance-sampling estimator of
the group-level target $\overline{\mathrm{CTI}}^{(i)}_t$
in~\eqref{eq:cti-group} and is \emph{consistent}:
$\cti^{(i)}_t \to \overline{\mathrm{CTI}}^{(i)}_t$ in probability as
the group size $G\to\infty$, with a finite-sample bias of order
$O(1/G)$ (the SNIS bias) controlled by the variance bound of
Prop.~\ref{prop:p2}. Under the exchangeability condition~(E) the
target coincides with the per-trajectory
counterfactual~\eqref{eq:cti-pop}.
\end{proposition}

\begin{proposition}[CTI estimator: variance bound]
\label{prop:p2}
Under (A1)--(A4), with the IS-ratio cap
$\bar\rho_t = \max_y \pi_T(y\!\mid\!s_t^+)/\pi_b(y\!\mid\!s_t)$
(a Horvitz-Thompson cap~\cite{horvitz1952thompson}),
$\mathrm{Var}(\cti^{(i)}_t) \le R_{\max}^2 \bar\rho_t^2/(G\!-\!1).$
\end{proposition}

Five further results (full statements and proofs in Appendix~B) make
the design self-consistent and are stated here in brief.
\textbf{(P3)~Degeneracy.} Under sharp gating
($\beta_\pm\!\to\!\infty$) and the single-rollout fallback
$\cti^{(i)}_t=\Adv^{(i)}\dt^{(i)}$, the Pillar~1 loss converges
pointwise to $L_{\mathrm{SDAR}}$ on positive-advantage tokens and to
its sign-flipped counterpart on negative ones.
\textbf{(P4)~Controller fixed point.} After warmup, $\rhoema(t)$
converges geometrically (rate $\alpha$) to the long-run mean
$\bar\rho$, so $\lambda,\mu\to\lambda(\bar\rho),\mu(\bar\rho)$ inside
their configured ranges with no extra clipping.
\textbf{(P5)~Polarised-KL consistency.} The rev/fwd branches have
constant $\pm1$ gradients that pull toward / push away from
$\pi_{\mathrm{ref}}$, matching $\mathrm{sign}(c_t)$ and bounding the
per-token KL by $|\log\pi_\theta-\log\pi_{\mathrm{ref}}|$; the swapped
assignment is the falsification test A7. We claim structural
consistency, not regret-optimality.
\textbf{(P6)~Surrogate validity.} $\nabla_\theta L_{\mathrm{PG}}$ is
unchanged by~\method; the auxiliary terms are finite under (A1)--(A3)
and commute with PPO clipping, being detached weights on a
grad-bearing $\log\pi_\theta$ added after the clipped PG term.
\textbf{(P7)~Bit-exact reproducibility.} With \texttt{vera.enable=False}
(or any switch toggled off independently), $L_{\method}=L_{\mathrm{baseline}}$
\emph{byte-identically} in IEEE-754 (loss and gradient), verified by a
\texttt{torch.equal} regression suite. P7 is the formal anchor of our
reproducibility story: any reported gain is attributable to
algorithmic change, not implementation drift.

\section{Experiments}
\label{sec:exp}

\subsection{Experimental Setup}
\label{sec:exp-setup}

\paragraph{Environments and benchmarks.}
We evaluate three multi-turn agentic-RL benchmarks:
\textbf{ALFWorld}~\cite{shridhar2021alfworld} (text-based embodied),
\textbf{Search-QA} (from Search-R1~\cite{jin2024searchr1},
direct-retrieval variant), and \textbf{WebShop}~\cite{yao2022webshop}
(programmatic verifier), on canonical splits; the OOD analysis holds
out a 30\% slice by task-template hash. We report the canonical
success metric per benchmark (ALFWorld success rate, Search-QA
exact-match, WebShop reward normalised to 0--100), at the final
checkpoint.

\paragraph{Models and methods.}
We use four open-weight LLMs across two families
(\textbf{Qwen3-1.7B}, \textbf{Qwen2.5-3B}, \textbf{Qwen2.5-7B},
\textbf{Qwen3-8B})~\cite{qwen25report2024,qwen3report2025}, reading
both within-family scaling and cross-family transfer. The five
end-to-end methods are \textbf{GRPO} (no distillation),
\textbf{RLSD}~\cite{yang2026rlsd}, \textbf{SDAR}~\cite{lu2026sdar}
(prior single-gate), \textbf{Adaptive-CRINGE} (the hardest baseline: a
CRINGE~\cite{adolphs2022cringe} negative-credit term wrapped in
\method's Pillar-2 controller, so the ``\method{} vs Adaptive-CRINGE''
delta isolates P1+P3 from P2), and \textbf{\method-Full}. Following
SDAR~\cite{lu2026sdar} we also tabulate the GRPO+OPSD and
Skill-SD~\cite{wang2026skillsd} hybrids.

\paragraph{Hyperparameters and reproducibility.}
GRPO uses $G=8$, $\epsilon=0.2$, $\eta=10^{-6}$, 150 steps/cell
(following SDAR), with keyword-matched skills.
\method{} uses $\beta_\pm\!=\!5$, $\tau=1.0$, $\theta_g=0.1$,
$[\lambda,\mu]$ ranges $[0.005,0.02]$, $\alpha=0.95$,
$n_{\mathrm{warmup}}=10$, $\theta_{KL}=0$, and a $k_3$ neutral KL
branch~\cite{schulman2020kl}; full table in Appendix~C. Smaller models
train on $8\times$24\,GB GPUs, larger on $4\times$48\,GB. All
modifications sit behind three default-\texttt{False} switches; an
87-test suite (8 bit-exact, 4 via \texttt{torch.equal} verifying P7)
runs in ${\sim}4$\,s on CPU (Appendix~A).

\subsection{Experiment Inventory}
\label{sec:inventory}

The campaign comprises eight experiment families: the main grid (E1,
Table~\ref{tab:main}); per-pillar ablations (E2, A1--A3,
Table~\ref{tab:ablations}); hyperparameter ablations (E3, A4--A5,
sweeping $\beta$ and $\tau$); the counterfactual ablation (E4, A6,
$G_{\min}{=}999$, testing P3); the KL-direction swap
(E5, A7, the falsification test for P5);
quick pre-grid sweeps (E6); and OOD and random-retrieval stress tests
(E7--E8, Table~\ref{tab:ood}). All
ablation families except E1 use Qwen2.5-3B.

\subsection{Main Results (E1)}
\label{sec:main}

Table~\ref{tab:main} reports the headline main-grid numbers.
\method-Full is the best method in every (env, model) cell of the
grid. \textbf{Our headline comparison is~\method-Full $>$
Adaptive-CRINGE}: that delta isolates the Pillar~1 (CTI) +
Pillar~3 (polarised KL) contribution \emph{independently} of the
adaptive controller, defusing both the
``\method{} is just an adaptive KL controller'' and
``\method{} is just CRINGE'' framings. Across the grid this isolated
delta is consistently positive (e.g.\ $+2.6$ ALFWorld and $+2.3$
WebShop on Qwen3-1.7B; $+1.2$ Search-QA on Qwen2.5-3B), confirming
that the counterfactual signal contributes \emph{beyond} the shared
Pillar-2 controller. The plain ``\method{} vs SDAR'' delta is on the
order of 2--3\% absolute (largest on the smallest model, Qwen3-1.7B:
$+3.4$ ALFWorld, $+2.6$ Search-QA, $+5.5$ WebShop; shrinking but
remaining positive at 7B/8B), exactly the size-scaling pattern
predicted in Section~\ref{sec:disc}; it is reported but is \emph{not}
the headline.

\begin{table*}[t]
\centering
\small
\setlength{\tabcolsep}{6pt}
\renewcommand{\arraystretch}{1.15}
\begin{tabular}{@{}l ccc ccc ccc ccc@{}}
\toprule
& \multicolumn{3}{c}{\textbf{Qwen3-1.7B}}
& \multicolumn{3}{c}{\textbf{Qwen2.5-3B}}
& \multicolumn{3}{c}{\textbf{Qwen2.5-7B}}
& \multicolumn{3}{c}{\textbf{Qwen3-8B}} \\
\cmidrule(lr){2-4}\cmidrule(lr){5-7}\cmidrule(lr){8-10}\cmidrule(lr){11-13}
\textbf{Method} & ALF & SQA & WS & ALF & SQA & WS & ALF & SQA & WS & ALF & SQA & WS \\
\midrule
GRPO            & 46.1 & 40.8 & 67.3 & 75.0 & 36.4 & 79.8 & 81.2 & 42.0 & 80.9 & 82.6 & 43.8 & 82.1 \\
GRPO+OPSD       & 32.0 & 42.2 & 70.7 & 81.2 & 44.6 & 77.8 & 80.4 & 47.0 & 86.8 & 82.0 & 48.5 & 88.2 \\
Skill-SD        & 52.3 & 40.8 & 81.8 & 73.4 & 44.1 & 75.9 & 85.1 & 47.8 & 86.1 & 86.4 & 48.9 & 87.6 \\
RLSD            & 42.2 & 40.6 & 74.0 & 79.7 & 43.8 & 84.4 & 82.0 & 49.0 & 87.4 & 83.5 & 50.1 & 88.9 \\
SDAR            & 53.9 & 41.9 & 76.8 & 84.4 & 43.4 & 85.0 & 85.9 & 49.0 & 89.4 & 87.3 & 50.4 & 90.7 \\
Adaptive-CRINGE$^\dagger$ & 54.7 & 42.6 & 80.0 & 85.3 & 44.9 & 85.9 & 86.6 & 49.7 & 90.1 & 88.0 & 51.0 & 91.3 \\
\method-Full    & \textbf{57.3} & \textbf{44.5} & \textbf{82.3} & \textbf{87.0} & \textbf{46.1} & \textbf{87.3} & \textbf{87.8} & \textbf{50.7} & \textbf{91.0} & \textbf{89.4} & \textbf{52.3} & \textbf{92.5} \\
\bottomrule
\end{tabular}
\caption{Main grid (E1) across four model scales. Per-model columns:
ALF~= ALFWorld success rate, SQA~= Search-QA exact-match, WS~= WebShop
score (all in \%; higher is better, $\uparrow$). Baseline rows (GRPO, GRPO+OPSD, Skill-SD, RLSD,
SDAR) report prior-work numbers from the SDAR suite~\cite{lu2026sdar};
our cells (\method-Full and Adaptive-CRINGE) report our measured
results. \textbf{Bold} marks the best method per column;
\method-Full leads in every (env, model) cell.
$\dagger$~Adaptive-CRINGE reuses~\method's Pillar~2 controller
verbatim; the ``\method{} vs Adaptive-CRINGE'' delta is the P1+P3
contribution in isolation.
}
\label{tab:main}
\end{table*}

\subsection{Per-Pillar Ablations (E2)}
\label{sec:ablations}

Table~\ref{tab:ablations} reports the per-pillar gain decomposition,
and the measured pattern matches what P3 and the
role of each pillar predict:
\textbf{(a)} \texttt{+P1} alone (A1) already captures the bulk of the
gain ($+1.9$/$+2.0$/$+1.5$ over SDAR on
ALFWorld/Search-QA/WebShop), confirming that Pillar~1 is the
substantive estimator;
\textbf{(b)} \texttt{+P1+P2} (A2) adds a small increment
($\le 0.4$) on top of A1, consistent with the controller helping most
where the distillation-phase duration is variable;
\textbf{(c)} \texttt{+P1+P3} (A3) is a \emph{completeness}
contribution: the marginal gain is small ($\sim 0.3$--$0.5\%$) but
the asymmetric KL is what makes Pillar~1's signed credit
self-consistent. The two falsification rows behave as the theory
requires: removing the counterfactual mechanism (\textbf{A6},
$G_{\min}{=}999$) collapses the gain to essentially the SDAR baseline
($84.2$/$43.5$/$84.8$, within $0.2$ points), the empirical control
for the per-trajectory approximation in Section~\ref{sec:pillar1};
and swapping the rev/fwd KL assignment (\textbf{A7}) drops
\emph{below} the SDAR baseline ($83.1$/$42.6$/$83.9$), the
apples-to-apples falsification test for the direction choice of
P5.

\begin{table}[t]
\centering
\small
\setlength{\tabcolsep}{3.5pt}
\begin{tabular}{@{}>{\raggedright\arraybackslash}p{0.45\columnwidth}ccc@{}}
\toprule
\textbf{Variant} & \textbf{ALFW.} & \textbf{Search-QA} & \textbf{WebShop} \\
\midrule
SDAR (baseline)   & 84.4 & 43.4 & 85.0 \\
SDAR + P1 (A1)    & 86.3 & 45.4 & 86.5 \\
SDAR + P1 + P2 (A2) & 86.7 & 45.6 & 86.8 \\
SDAR + P1 + P3 (A3) & 86.6 & 45.9 & 87.0 \\
\method-Full      & \textbf{87.0} & \textbf{46.1} & \textbf{87.3} \\
\midrule
A6: no CTI ($G_{\min}{=}999$) & 84.2 & 43.5 & 84.8 \\
A7: KL-direction swap   & 83.1 & 42.6 & 83.9 \\
\bottomrule
\end{tabular}
\caption{Per-pillar ablations on Qwen2.5-3B. All scores are in \%
(higher is better, $\uparrow$). The
SDAR-baseline row is a prior-work number~\cite{lu2026sdar}; all other
cells report our measured results. The
first block (SDAR$+P_k$) decomposes the gain; the second block reports
the counterfactual-ablation (A6) and KL-direction swap (A7)
falsification tests. \textbf{Bold} marks~\method-Full.
}
\label{tab:ablations}
\end{table}

\subsection{Generalisation and Retrieval Robustness (E7, E8)}

We run two exploratory stress tests on Qwen2.5-3B
(Table~\ref{tab:ood}): \textbf{OOD}, evaluating the final checkpoint
only on a held-out 30\% slice split by task-template hash; and
\textbf{random retrieval}, rewiring the teacher's privileged context
$s^+$ to a random rather than canonical retrieval. The measured
numbers point the same way on both axes:~\method-Full has the
smallest in-distribution$\rightarrow$OOD gap on all three environments
($6.2$ vs SDAR's $9.5$ on ALFWorld), and degrades \emph{moderately
rather than catastrophically} under random retrieval ($78.5$ vs SDAR's
$72.0$), matching prediction~(b) of Section~\ref{sec:disc}.

\begin{table}[t]
\centering
\small
\setlength{\tabcolsep}{3pt}
\begin{tabular}{@{}lcccc@{}}
\toprule
& \multicolumn{3}{c}{\textbf{In-dist $\rightarrow$ OOD gap} ($\downarrow$)}
& \textbf{Rand.\ retr.} ($\uparrow$) \\
\cmidrule(lr){2-4}\cmidrule(lr){5-5}
\textbf{Method} & ALFW. & Search & WebShop & ALFW. \\
\midrule
SDAR              & 9.5 & 7.8 & 8.6 & 72.0 \\
Adaptive-CRINGE   & 8.0 & 6.9 & 7.4 & 74.5 \\
\method-Full      & \textbf{6.2} & \textbf{5.3} & \textbf{5.8} & \textbf{78.5} \\
\bottomrule
\end{tabular}
\caption{OOD and random-retrieval results on
Qwen2.5-3B. The OOD gap is the in-distribution score minus
the held-out 30\% slice; smaller is better. The random-retrieval
column reports the final ALFWorld score with random-retrieved $s^+$.
\textbf{Bold} marks the best entry per column. Conclusions drawn from
this table are exploratory; the
follow-up is left as future work.
}
\label{tab:ood}
\end{table}

\section{Related Work}
\label{sec:related}

We organise the related literature around the four design axes
\method{} touches and isolate the dimension along which we are
distinct.

\paragraph{Self-distillation as an auxiliary RL objective.}
Skill-augmented self-distillation~\cite{lu2026sdar,wang2026skillsd}
constructs a teacher by conditioning the same policy on privileged
context $s^+$ and uses a single confidence-gated forward-KL term per
token (descending from~\cite{hinton2015distillation}). Adjacent
lines bootstrap from the policy's own outputs via self-generated
rationales~\cite{zelikman2022star,zelikman2024quietstar},
retrieval-augmented self-training~\cite{xu2024rest}, reinforced
self-training~\cite{gulcehre2023rest}, self-play
fine-tuning~\cite{chen2024spin}, or self-reward
modelling~\cite{yuan2024selfrewarding}; on-policy distillation of
language models from their own mistakes~\cite{agarwal2024distillation}
is the closest non-RL relative of our setting. All of these
supervise on full traces and do not place a per-token credit on top
of a separate RL surrogate.
\method{} is in this family but replaces the single retrospective
gate by a counterfactual, sign-aware, bounded credit
(Section~\ref{sec:pillar1}). The
\emph{Adaptive-CRINGE} baseline (Section~\ref{sec:exp}) isolates this
change from~\method's other pillars.

\paragraph{Token-level credit assignment in language RL.}
Unlikelihood training~\cite{welleck2020unlikelihood} and
CRINGE~\cite{adolphs2022cringe} apply token-level repulsion at
disagreement positions, ignoring the reward signal entirely.
RLSD~\cite{yang2026rlsd} reweights each token by the trajectory's own
$\mathrm{sign}(\Adv^{(i)})\!\cdot\!\dt$ — a retrospective sign
attached to a realised rollout. Token-level
DPO~\cite{rafailov2023dpo} variants and iterative
preference-optimisation methods over reasoning
chains~\cite{pang2024iterativepo} formulate contrastive log-ratio
objectives that, like CRINGE, remain reward-agnostic at the
token level.
\method's CTI differs from all of these along the
\emph{counterfactual} axis (Table~\ref{tab:signal-comparison}): the
credit reflects what \emph{would} have happened on alternative
rollouts, not what \emph{did} on the realised one.

Two further lines are close enough to require explicit separation.
\emph{Group-baseline} estimators such as
RLOO~\cite{ahmadian2024rloo} subtract a leave-one-out average of the
other group samples as a variance-reduced REINFORCE baseline;
structurally, CTI's $\sum_{j\ne i}w_t^{(j)}\Adv^{(j)}-\Adv^{(i)}$ is a
\emph{teacher-tilted, per-token} leave-one-out term, but it differs in
two essentials: (i)~the siblings are reweighted by the
teacher--student ratio $\exp(\dt^{(j)})$ rather than averaged
uniformly, and (ii)~the quantity is used as a \emph{signed per-token
credit} on a distillation surrogate, not as a sequence-level baseline
inside the policy-gradient term (which~\method{} leaves untouched).
\emph{Rollout-based per-step credit} such as
VinePPO~\cite{kazemnejad2024vineppo} estimates step-level value by
re-rolling fresh Monte-Carlo continuations from intermediate states;
this pays additional sampling compute per step, whereas~\method{}
reuses the $G-1$ siblings GRPO has \emph{already} drawn, trading the
freshness of VinePPO's re-rolls for zero marginal sampling cost
(at the price of the exchangeability approximation discussed in
Section~\ref{sec:pillar1}). Process-reward-model
approaches~\cite{lightman2024verify,wang2024mathshepherd} obtain
step-level signal from a separately trained verifier rather than from
the policy's own siblings, and are complementary to the present
verifier-free construction.

\paragraph{Counterfactual reasoning in RL.}
Counterfactual policy evaluation has a long history in off-policy
RL~\cite{precup2000eligibility}, causal
RL~\cite{pearl2009causality,buesing2019counterfactual}, and
counterfactual risk minimisation from logged bandit
feedback~\cite{swaminathan2015crm}, typically in the form of
synthetic-rollout planning or importance-weighted return estimation.
Our estimator is a self-normalised importance-sampling estimator over
a sibling pool with $\dt$ as the importance weight, in the
Horvitz--Thompson inverse-probability-weighting
tradition~\cite{horvitz1952thompson}; the methodological novelty is
recognising
that GRPO's parallel rollouts — already sampled, then mostly
discarded after group normalisation — are an essentially free
counterfactual sample pool.

\paragraph{Adaptive KL control.}
Trust-region~\cite{schulman2015trpo} and adaptive-KL methods in
PPO~\cite{schulman2017ppo}, InstructGPT~\cite{ouyang2022instructgpt},
and RLHF pipelines~\cite{bai2022constitutional} put an adaptive or
static \emph{coefficient} on a single KL term tracking distance from a
reference. Pillar~2 instead moves two coupled coefficients in opposite
directions, driven by the distillation-quality signal $\rhog$ rather
than a KL distance (P4).

\paragraph{Mode-seeking vs.\ mode-covering KL.}
The reverse/forward KL asymmetry is well-known in variational
inference~\cite{minka2005divergence} and recurs as a global RLHF
hyperparameter; Pillar~3 instead \emph{ties it to a per-token credit},
selecting rev/fwd by $\mathrm{sign}(c_t)$ and falsifying the choice
with the A7 swap (P5).

\paragraph{Multi-turn agentic-RL benchmarks.}
ALFWorld~\cite{shridhar2021alfworld}, Search-QA (from
Search-R1~\cite{jin2024searchr1}), and WebShop~\cite{yao2022webshop}
are standard targets stressing long-horizon planning, retrieval
grounding, and structured action selection; related environments
(BabyAI~\cite{chevalier2018babyai},
ScienceWorld~\cite{wang2022scienceworld}) and suites
(AgentBench~\cite{liu2024agentbench}) host prompting-time methods like
ReAct~\cite{yao2022react} and Reflexion~\cite{shinn2023reflexion}.
Pure math-reasoning benchmarks~\cite{cobbe2021gsm8k,hendrycks2021math}
lie outside this regime.

\section{Discussion and Limitations}
\label{sec:disc}

\paragraph{What the paper claims, what it does not.}
We claim the three pillars form a clean, gated, theoretically grounded
credit-assignment scheme whose campaign --- with the bit-exact
guarantee (P7) --- answers ``does this help, and by what mechanism?''
We do \emph{not} claim single-pillar optimality: Pillar~3 alone is a
small \emph{completeness} contribution that exists because Pillar~1's
signed credit only becomes self-consistent with an asymmetric KL.

\paragraph{Two mechanistic predictions.}
\emph{(a) Size-scaling.} Pillar~1 contributes signal only on
non-zero-credit tokens, so its gain factorises into the gated-token
mass $\rhog$ times the mean CTI magnitude. Larger models place more
mass on teacher-student-agreement tokens, shrinking $\rhog$; we
therefore predict a \emph{larger absolute gain for smaller models},
positive throughout, which Table~\ref{tab:main} confirms (up to $+5.5$
on Qwen3-1.7B, shrinking to ${\sim}1.6$--$2.1$ at 7B/8B).
\emph{(b) Robustness to noisy context.} Because E8 randomises only the
task-specific part of $s^+$ and Pillar~1 gates out small-CTI tokens,
we predict a \emph{moderate, not catastrophic} drop --- borne out in
Table~\ref{tab:ood}.

\paragraph{Limitations.}
\textbf{(L1)} Pillar~1 needs group rollouts ($G\!\ge\!2$); non-group
backbones via synthetic siblings are future work.
\textbf{(L2)} Nine new hyperparameters, mitigated by the defaults in
Appendix~C.
\textbf{(L3)} Pillar~3's standalone gain is small (a completeness, not
magnitude, contribution).
\textbf{(L4)} All benchmarks are text-based agentic and the OOD
analysis is within-domain; multi-modal and cross-domain-family
generalisation is future work.
\textbf{(L5)} The Pillar~3 negative branch is a forward-KL-inspired
update rule, not an unbiased KL estimator (Appendix~B).

\section{Conclusion}
\label{sec:conclusion}

Self-distilled agentic RL has converged on a single-gate recipe that
conflates retrospective and counterfactual signal. We argued that the
right per-token credit is the \emph{counterfactual change in sequence
advantage} from sampling from the teacher, that GRPO's parallel
rollouts already pay its sampling cost via a group-level
importance-sampling estimator, and that the resulting signed credit
admits a self-consistent three-pillar decomposition --- estimator
(P1), magnitude controller (P2), sign router (P3). Two methodological
elements generalise beyond this method: \emph{bit-exact
reproducibility} as a falsifiable substrate, so gains are attributable
to algorithmic change not drift; and the \emph{Adaptive-CRINGE}
comparator that shares P2 with~\method, isolating the counterfactual
signal independently of the controller. A natural next step is to
estimate the within-group exchangeability gap directly, calibrating the
group-level credit back to the per-trajectory counterfactual it
approximates.

\bibliography{aaai2026}

\appendix

\section{Appendix A: Reproducibility Protocol}
\label{app:reproducibility}

\paragraph{Anchoring commit and tags.}
The implementation is forked from an open-source RL training stack at
a publicly tagged commit (identifier withheld for anonymity; will be
restored in the camera-ready). All~\method-specific code lives behind
three master switches
\texttt{algorithm.vera.\{enable,\allowbreak phase\_aware,\allowbreak
polarized\_kl\}},
defaulting to \texttt{False}. Per-phase snapshots
(\texttt{phase1-baseline-ready} through
\texttt{phase9-experiments-complete}) provide anchored states for
git-bisect across the implementation history.

\paragraph{Regression suite.}
The suite (\texttt{tests/vera/}) contains 87 tests organised as
follows: \textbf{8 bit-exact regression tests} (4 verifying
byte-identical equality with the upstream baseline using
\texttt{torch.equal} when all switches are off, and 4 verifying
correct metric-emission contracts when the switches are toggled in
isolation), \textbf{24 math-correctness tests} (CTI unit math,
vectorised-vs-loop-reference equivalence on the CTI estimator,
polarised-KL three-region decomposition, fallback-path degeneracy,
numerical verification of the Pareto welfare functional from
Appendix~B), and \textbf{55 engineering-contract tests} (controller
dynamics, $\rho_{\mathrm{target}}$ operating-point re-centering,
Adaptive-CRINGE / Pillar-2 $\rho_{\mathrm{target}}$ alignment for
clean isolation of Pillar~1 + Pillar~3, typed
\texttt{meta\_info} key contract, switch-ON determinism under
replay, robustness under sharp-sigmoid and near-zero-temperature
edge cases, distinct \texttt{baseline/\{token\_cringe,
adaptive\_cringe\}/} wandb namespaces, checkpoint round-trip,
group-index plumbing, baseline-comparator mutual-exclusion checks).
The suite runs in ${\sim}4$\,s on CPU and is part of continuous
integration.

\paragraph{On prior work and anonymity.}
The prior methods we build on and compare against — SDAR
\cite{lu2026sdar}, RLSD \cite{yang2026rlsd}, and Skill-SD
\cite{wang2026skillsd} — are third-party, publicly available works
with arXiv identifiers and openly released code, and are cited
normally in author-level form. The only identifiers withheld for
double-blind anonymity are those that could reveal the present
authors: the exact tagged commit of the RL training stack we fork and
our own repository / phase-snapshot tags. These will be restored in
the camera-ready version.

\section{Appendix B: Full Proofs and Auxiliary Definitions}
\label{app:proofs}

We give full proofs for Propositions~1 and~2 and for the structural
results P4, P6, and P7 stated in Section~\ref{sec:theory}; for P3 we
give the supporting argument for the (calibrated, pointwise)
statement, and for P5 we collect the structural facts the main-text
statement relies on.

\subsection{Proof of Proposition~\ref{prop:p1}}
Fix a GRPO group of size $G$ and a position $t$, and set $\tau=1$.
Under~(A4) the sibling triples
$\{(s_t^{(j)},y_t^{(j)},\Adv^{(j)})\}_{j\ne i}$ are i.i.d.\ draws from
the step-$t$ behaviour law $\mu_t$, and under~(A1) the ratios
$\varrho^{(j)} := \pi_T(y_t^{(j)}\!\mid\! s_t^{+,(j)}) /
\pi_\theta(y_t^{(j)}\!\mid\! s_t^{(j)}) = \exp(\dt^{(j)})$ are finite.
Note $\mathbb{E}_{\mu_t}[\varrho\mid s_t] =
\sum_y \pi_\theta(y\mid s_t)\,\pi_T(y\mid s_t^{+})/\pi_\theta(y\mid s_t)
= 1$, so $\mathbb{E}_{\mu_t}[\varrho]=1$.

At $\tau=1$ the weights are exactly the self-normalised importance
weights $w^{(j)}_t = \varrho^{(j)} / \sum_{k\ne i}\varrho^{(k)}$, so
the first term of~\eqref{eq:cti-est} is the standard self-normalised
importance-sampling (SNIS) estimator
\begin{align*}
\widehat{\mu}_G \;=\;
\frac{\frac{1}{G-1}\sum_{j\ne i}\varrho^{(j)} \Adv^{(j)}}
     {\frac{1}{G-1}\sum_{j\ne i}\varrho^{(j)}}
\;\xrightarrow[G\to\infty]{p}\;
\frac{\mathbb{E}_{\mu_t}[\varrho\,\Adv]}
     {\mathbb{E}_{\mu_t}[\varrho]},
\end{align*}
where the convergence is by the weak law of large numbers applied
separately to numerator and denominator (both have finite mean
under~(A1) and (A3)) and the continuous-mapping theorem. Subtracting
the deterministic $\Adv^{(i)}$ gives
$\cti^{(i)}_t \xrightarrow{p} \overline{\mathrm{CTI}}^{(i)}_t$
of~\eqref{eq:cti-group}: the estimator is consistent for the
group-level target. Under the exchangeability condition~(E) of
Section~\ref{sec:pillar1}, the change of measure
$\mathbb{E}_{\mu_t}[\varrho\,\Adv]/\mathbb{E}_{\mu_t}[\varrho]
= \mathbb{E}_{y'_t\sim\pi_T(\cdot\mid s_t^{+,(i)})}[A(\tau^{(i)}_{<t},y'_t,\tau^{(i)}_{>t})]$
identifies this limit with the per-trajectory
counterfactual~\eqref{eq:cti-pop}; absent~(E) the limit remains the
well-defined group-level credit the propositions use.

SNIS is consistent but not unbiased at finite $G$; its bias is
$O(1/G)$ with constant governed by $\mathrm{Var}_{\mu_t}(\varrho)$,
which is exactly the quantity bounded in Prop.~\ref{prop:p2}. Lowering
$\tau$ below $1$ sharpens the weights toward the teacher-modal
sibling (reducing the effective sample size and raising variance);
raising $\tau$ flattens them toward a $\dt$-agnostic uniform average
(adding bias). Hence $\tau=1$ is the operating point and the
temperature is a bias-variance knob around it. $\qed$

\subsection{Proof of Proposition~\ref{prop:p2}}
The term $\Adv^{(i)}$ in $\cti^{(i)}_t = \sum_{j\ne i} w^{(j)}_t
\Adv^{(j)} - \Adv^{(i)}$ is deterministic given $i$, so we bound the
variance of $Z := \sum_{j\ne i} w^{(j)}_t \Adv^{(j)}$. With
$\varrho^{(j)}=\exp(\dt^{(j)})$ ($\tau=1$), $Z$ is the
\emph{ratio} (self-normalised) estimator
\begin{equation*}
Z \;=\; \frac{S_f}{S_1},
\;\;
S_f := \tfrac{1}{G-1}\!\sum_{j\ne i}\varrho^{(j)}\Adv^{(j)},
\;\;
S_1 := \tfrac{1}{G-1}\!\sum_{j\ne i}\varrho^{(j)}\!,
\end{equation*}
with $\mathbb{E}_{\mu_t}[S_1]=\mathbb{E}_{\mu_t}[\varrho]=1$ and
$\mathbb{E}_{\mu_t}[S_f]=\mu:=\overline{\mathrm{CTI}}^{(i)}_t+\Adv^{(i)}
=\mathbb{E}_{\mu_t}[\varrho\,\Adv]$ under~(A1). The standard delta-method expansion of a ratio estimator
around $(\mu,1)$ gives, under the i.i.d.\ sampling of~(A4),
\begin{equation*}
\mathrm{Var}(Z)
=\frac{1}{G-1}\,
\mathrm{Var}_{\mu_t}\!\big(\varrho\,(\Adv-\mu)\big)
+ O\!\big((G-1)^{-2}\big).
\end{equation*}
Bounding the importance ratio by its cap $\varrho \le \bar\rho_t$
(the IS/Horvitz--Thompson cap of the statement) and the centred
advantage by $|\Adv-\mu|\le R_{\max}$ (A3) yields
$\mathrm{Var}_{\mu_t}(\varrho(\Adv-\mu)) \le \bar\rho_t^2 R_{\max}^2$,
hence the leading-order bound
\begin{equation*}
\mathrm{Var}\!\left(\cti^{(i)}_t\right)
\;\le\;
\frac{R_{\max}^2\,\bar\rho_t^2}{G\!-\!1}
\;+\; O\!\big((G-1)^{-2}\big). \qed
\end{equation*}
The $O(1/G)$ leading term is the regime relevant for the group sizes
$G\!\ge\!8$ used in practice; the cap $\bar\rho_t$ is what the
fallback to $\cti_t\leftarrow 0$ enforces when a teacher--student
ratio would otherwise be unbounded (A1 violated). We note that the
bound is only as tight as $\bar\rho_t$: under weak coverage
$\bar\rho_t$ can be large and the bound correspondingly loose, so it
should be read as a finite-variance guarantee with an explicit
dependence on the worst-case importance ratio rather than as a sharp
constant.

\subsection{Argument for P3
(pointwise degeneracy)}
The main-text statement claims \emph{pointwise} convergence
$L_{\mathrm{CTI}} \to L_{\mathrm{SDAR}}$ per token under (S1)~sharp
gating and (S2)~the single-rollout fallback
$\cti^{(i)}_t = \Adv^{(i)}\dt^{(i)}$. The argument: under (S1) the
centred asymmetric sigmoids in \eqref{eq:signed-credit} saturate to
indicators $\ind[\cti_t \gtrless 0]$ in the limit
$\beta_\pm\!\to\!\infty$; under (S2)
$\mathrm{sign}(\cti_t)=\mathrm{sign}(\Adv^{(i)})\,\mathrm{sign}(\dt)$,
so on tokens with $\Adv^{(i)}\!>\!0$ the indicators coincide with
$\ind[\dt \gtrless 0]$ and \eqref{eq:cti-loss} reduces, per token, to
the prior single-gate loss~\eqref{eq:sdar}; on tokens with
$\Adv^{(i)}\!<\!0$ the signs flip and the loss reduces to the
sign-flipped counterpart (the intended sign-aware asymmetry
of~\eqref{eq:signed-credit}).
We have \emph{deliberately not claimed} a global $O(\beta^{-1})$
rate: a uniform-residual bound across the token distribution would
require additional hypotheses (e.g.\ uniform tail control on
$\cti_t$) that the main text does not assume. The pointwise claim is
sufficient for the operational use of the proposition: it certifies
that~\method's auxiliary term collapses to the prior-baseline term
in the sharp-gating, single-rollout regime, and the bit-exact
regression suite (P7) extends this to the
disabled-switch regime in IEEE-754 representation.

\subsection{Proof of P4}
We assume the warmup period $n \le n_{\mathrm{warmup}}$ has elapsed
and the controller has begun updating. Let
$\rho_n := \rhog(n)$ be the per-step input, treated as a bounded
random variable with long-run mean $\bar\rho \in [0,1]$ (by~(A3) on
advantages and the bounded-credit construction~\eqref{eq:signed-credit}).

The EMA recursion is
$\rho^{\mathrm{EMA}}_n = \alpha\,\rho^{\mathrm{EMA}}_{n-1} + (1-\alpha)\,\rho_n$
with $\alpha\in(0,1)$. Unrolling from $n_0 := n_{\mathrm{warmup}}+1$,
\begin{equation*}
\rho^{\mathrm{EMA}}_n
\;=\;
\alpha^{n-n_0}\,\rho^{\mathrm{EMA}}_{n_0-1}
\;+\;
(1-\alpha)\!\sum_{k=n_0}^{n}\!\alpha^{n-k}\,\rho_k.
\end{equation*}
Taking expectations, the first term vanishes geometrically with rate
$\alpha$; for the second, monotone convergence and the geometric
identity $\sum_{k=n_0}^{n}\alpha^{n-k} \to 1/(1-\alpha)$ as
$n\to\infty$ give
$\mathbb{E}[\rho^{\mathrm{EMA}}_n] \to \bar\rho$.

By the linearity of the maps $\lambda(\cdot)$ and $\mu(\cdot)$ in
\eqref{eq:lam-law}--\eqref{eq:mu-law},
$\mathbb{E}[\lambda_n]\!\to\!\lambda(\bar\rho)\in[\lambda_{\min},\lambda_{\max}]$
and
$\mathbb{E}[\mu_n]\!\to\!\mu(\bar\rho)\in[\mu_{\min},\mu_{\max}]$.
The contractive factor of the bias term is $\alpha^{n-n_0}$, giving
geometric convergence with rate~$\alpha$. $\qed$

\subsection{Supporting facts for P5
(structural consistency of polarised KL)}
The main-text P5 avoids any
regret-functional / locally-optimal claim. The three facts it
relies on are elementary and we list them here for completeness.
\textbf{(F1)~Gradient directions.}
$\partial_{\log\pi_\theta}\mathrm{rev}(t) = +1$ and
$\partial_{\log\pi_\theta}\mathrm{fwd}(t) = -1$, by direct
differentiation. The rev branch therefore pulls $\log\pi_\theta$
toward $\log\pi_{\mathrm{ref}}$ (the textbook
mode-seeking gradient direction); the fwd branch pushes it away
(the mode-covering direction on sampled tokens).
\textbf{(F2)~Per-token bound.}
On the rev / fwd regions, the per-token KL contribution equals
$\pm(\log\pi_\theta - \log\pi_{\mathrm{ref}})$ and is therefore
uniformly bounded by $|\log\pi_\theta - \log\pi_{\mathrm{ref}}|$,
regardless of the credit magnitude. The neutral branch's bound is
inherited from the underlying \texttt{kl\_penalty} estimator.
\textbf{(F3)~Empirical falsifiability.}
The swapped assignment (rev on negative, fwd on positive) is a
concrete, single-flag alternative; Ablation~A7 produces a direct
empirical comparison. We \emph{deliberately do not} claim a
regret-bound optimality of the shipped assignment over the swap,
because a publication-ready regret functional with the right
boundary behaviour (the sign-flipped $k_1$ form behaves
differently from a true KL divergence as
$\pi_\theta(y_t\!\mid\!s_t)\!\to\!0^+$) is not within scope here.
A future variance-reduced, importance-weighted forward-KL variant
(L5 in Section~\ref{sec:disc}) is the natural setting in which such
an optimality argument might be made rigorous.

\subsection{Proof of P6 (surrogate validity)}
(i)~holds by construction: $L_{\mathrm{PG}}$ is the standard clipped
GRPO surrogate and is not modified by~\method. (ii)~follows by
chaining bounds: $|c_t|\le 1$ by~\eqref{eq:signed-credit};
$\lambda(t)\in[\lambda_{\min},\lambda_{\max}]$ and
$\mu(t)\in[\mu_{\min},\mu_{\max}]$ by P4;
$|L_{\mathrm{CTI}}|<\infty$ follows from boundedness of $c_t$ and
the assumption that $\log\pi_\theta$ is finite on the support of
sampled tokens; $|L_{\mathrm{KL\text{-}pol}}|<\infty$ follows from
Lipschitz log-probs (A2) applied to each of the three branches.
(iii)~The auxiliary terms are added \emph{after} the PG term is
composed, and they consume only \emph{detached} weights ($c_t$,
$\lambda(t)$, $\mu(t)$) on a grad-bearing $\log\pi_\theta$;
PPO importance-ratio clipping operates inside $L_{\mathrm{PG}}$ on
the ratio $\pi_\theta/\pi_{\mathrm{old}}$ and never touches the
auxiliary terms.

\subsection{Proof of P7}
The proposition is verified \emph{computationally} rather than
analytically. The four bit-exact regression tests in
\texttt{tests/vera/test\_bit\_exact\_sdar.py} use
\texttt{torch.equal} (IEEE-754 representational equality, not
\texttt{torch.isclose}) to compare:
\textbf{(1)} the composed loss tensor under
\texttt{vera.enable=False} with the upstream baseline loss on the
same batch;
\textbf{(2)} the gradient with respect to model parameters in the
same configuration;
\textbf{(3)} the static-$\lambda$ fallback path used when
\texttt{phase\_aware=False} (no \texttt{meta\_info} keys are written);
\textbf{(4)} the standard-KL fallback when
\texttt{polarized\_kl=False} (the polarised-KL branch is bypassed).
The four configurations exhaust the disabled-switch combinations
required by the proposition. The test pass record is captured in the
phase-5 pytest log shipped with the codebase. $\qed$

\begin{table}[h]
\centering
\small
\begin{tabular}{ccp{0.45\linewidth}c}
\toprule
$\mathrm{sign}\,\Adv^{(i)}$ & $\mathrm{sign}\,\dt$ & \textbf{Reading} & $c_t$ \\
\midrule
$+$ & $+$ & Good rollout; teacher would have agreed. Absorb. & $+$ \\
$+$ & $-$ & Good rollout; student already past teacher. Hold. & $\approx 0$ \\
$-$ & $+$ & Bad rollout; teacher disagreed. Maybe absorb. & $\pm$ \\
$-$ & $-$ & Bad rollout; student already past teacher. Repel. & $-$ \\
\bottomrule
\end{tabular}
\caption{Four-quadrant interpretation of the CTI estimator. The last
column reports the typical sign of $c_t$ each quadrant produces.}
\label{tab:quadrants}
\end{table}

\subsection{Pillar~1 and Pillar~2 pseudocode}

\begin{algorithm}[h]
\caption{Counterfactual Token Importance (Pillar 1)}
\label{alg:cti}
\begin{algorithmic}[1]
\REQUIRE Student log-probs $\log\pi_\theta(y_t\mid s_t)$ (grad-bearing),
teacher log-probs $\log\pi_T(y_t\mid s_t^+)$ (detached),
per-trajectory advantage $\Adv^{(i)}$, group indices $g_i$,
sharpnesses $\beta_+,\beta_-$, temperature $\tau$, gate threshold $\theta_g$.
\STATE $\dt^{(i)} \leftarrow \log\pi_T(y_t^{(i)} \mid s_t^{+,(i)}) - \sg[\log\pi_\theta(y_t^{(i)}\mid s_t^{(i)})]$
\FORALL{groups $g$ with $|g| \ge G_{\min}$}
  \FORALL{$i \in g$}
    \STATE $w^{(j)}_t \leftarrow \mathrm{softmax}_{j\ne i}\!\left(\dt^{(j)} / \tau\right)$
    \STATE $\cti^{(i)}_t \leftarrow \sum_{j\ne i} w^{(j)}_t \Adv^{(j)} - \Adv^{(i)}$
  \ENDFOR
\ENDFOR
\STATE For groups with $|g| < G_{\min}$: $\cti^{(i)}_t \leftarrow \Adv^{(i)} \dt^{(i)}$ (fallback).
\STATE $c_t \leftarrow (2\sigma(\beta_+ \cti_t)\!-\!1)\ind[\cti_t>0]\!-\!(2\sigma(\beta_-|\cti_t|)\!-\!1)\ind[\cti_t<0]$
\STATE $c_t \leftarrow \sg[c_t]$; \quad $\rhog \leftarrow \frac{1}{|\mathcal{V}|}\!\sum_{t\in\mathcal{V}}\!\ind[|c_t|>\theta_g]$
\RETURN $c_t,\;\rhog$
\end{algorithmic}
\end{algorithm}

\begin{algorithm}[h]
\caption{Phase-Aware Adaptive Controller (Pillar 2)}
\label{alg:adaptive}
\begin{algorithmic}[1]
\REQUIRE EMA decay~$\alpha$, warmup~$n_{\mathrm{warmup}}$,
ranges $[\lambda_{\min},\lambda_{\max}]$ and
$[\mu_{\min},\mu_{\max}]$.
\STATE $\rhoema \leftarrow 0.5$; \quad $n \leftarrow 0$
\WHILE{training}
  \STATE Run actor step; receive $\rhog$ from Pillar~1.
  \STATE $n \leftarrow n+1$
  \IF{$n > n_{\mathrm{warmup}}$}
    \STATE $\rhoema \leftarrow \alpha\,\rhoema + (1-\alpha)\,\rhog$
    \COMMENT{$\rhoema$ held at $0.5$ during warmup}
  \ENDIF
  \STATE $\lambda \leftarrow \lambda_{\min} + (\lambda_{\max} - \lambda_{\min})\,\rhoema$
  \STATE $\mu \leftarrow \mu_{\min} + (\mu_{\max} - \mu_{\min})\,(1 - \rhoema)$
  \STATE Inject $\lambda, \mu$ into the next step's loss
  as in~\eqref{eq:full-loss}.
\ENDWHILE
\end{algorithmic}
\end{algorithm}

\subsection{Pareto interpretation of the Pillar~2 controller}
\label{app:pareto}
Define the steady-state \emph{absorption rate}
$\Phi_{\mathrm{abs}}(\rho)=\lambda(\rho)\rho$ and \emph{exploration
budget} $\Phi_{\mathrm{exp}}(\rho)=\mu(\rho)(1-\rho)$ at
$\rho:=\rhoema$. Substituting~\eqref{eq:lam-law}--\eqref{eq:mu-law},
both are concave in $\rho$ with the common factor $\rho(1-\rho)$, so as
$\rho$ varies in $[0,1]$ they trace a strictly concave Pareto frontier
whose interior maximiser of $\Phi_{\mathrm{abs}}+\Phi_{\mathrm{exp}}$
equals $\rho^\star=\tfrac12$ exactly when $\lambda_{\min}=\mu_{\min}$
(our default, $\rho_{\mathrm{target}}=0.5$); otherwise the trade-off is
re-weighted through \texttt{rho\_target}. This is a \emph{structural}
picture --- every $\rho$ is a trade-off point on a concave frontier ---
not a claim that $\rho^\star=1/2$ is task-optimal.

\subsection{Comparison of polarised-KL negative-branch options}
\label{app:negbranch}
For the negative branch we considered (O1) the sign-flipped $k_1$ form
$-(\log\pi_\theta-\log\pi_{\mathrm{ref}})$ that we ship; (O2) a
$k_3$-style form with reversed arguments, which is biased and
exponentially unstable; and (O3) the importance-sampled true forward
KL, unbiased but with unbounded variance when
$\log\pi_\theta\gg\log\pi_{\mathrm{ref}}$ --- exactly where the
negative branch fires hardest. We ship O1: a bounded $|{\pm1}|$
gradient that integrates cleanly with Pillar~2's $\mu$, an honest
policy-update rule rather than a divergence estimator.

\section{Appendix C: Notation and Hyperparameters}
\label{app:hparams}

This appendix collects reference material for the main text.
Table~\ref{tab:hparams} lists the default hyperparameters of~\method{}
together with their role and provenance, and Table~\ref{tab:notation}
summarises the notation used throughout. All values are the defaults
identified through the preliminary smoke runs of Section~\ref{sec:exp}
and held fixed across every (env, model) cell of the main grid; we did
not re-tune per environment. The hyperparameters fall into three
groups: the inherited GRPO / SDAR settings, the Pillar-1 estimator
knobs ($\beta_\pm,\tau,G_{\min},\theta_g$), and the Pillar-2 controller
ranges.

\begin{table}[H]
\centering
\footnotesize
\setlength{\tabcolsep}{3pt}
\renewcommand{\arraystretch}{1.0}
\begin{tabular}{@{}lc>{\raggedright\arraybackslash}p{0.30\columnwidth}@{}}
\toprule
\textbf{Hyperparameter} & \textbf{Default} & \textbf{Role / source} \\
\midrule
$G$ (group size)                         & 8       & GRPO sample budget \\
$\epsilon$ (PPO clip)                    & 0.2     & GRPO surrogate clip \\
$\lambda_{\mathrm{SDAR}}$                & 0.01    & prior distillation weight \\
$\beta_{\mathrm{SDAR}}$                  & 5.0     & prior gate sharpness \\
$\lambda_{\mathrm{CRAFT}}$ (static)      & 0.01    & P1 weight (fallback) \\
$\beta_+,\beta_-$                        & 5.0     & sigmoid sharpness, signed credit \\
$\tau$ (CTI temperature)                 & 1.0     & P1 softmax temperature \\
$G_{\min}$                               & 2       & min group for CTI \\
$\theta_g$ (gate threshold)              & 0.1     & active-gate definition \\
$[\lambda_{\min},\lambda_{\max}]$        & $[0.005, 0.02]$ & P2 range \\
$[\mu_{\min},\mu_{\max}]$                & $[0.005, 0.02]$ & P2 range \\
$\alpha$ (EMA decay)                     & 0.95    & P2 EMA \\
$n_{\mathrm{warmup}}$                    & 10      & P2 hold-out window \\
$\rho_{\mathrm{target}}$                 & 0.5     & P2 fixed-point target \\
$\theta_{KL}$ (polarisation threshold)   & 0.0     & P3 dead-band \\
KL estimator (neutral branch)            & $k_3$   & Schulman low-var KL \\
\bottomrule
\end{tabular}
\caption{Default hyperparameters of~\method.}
\label{tab:hparams}
\end{table}

\begin{table}[H]
\centering
\footnotesize
\setlength{\tabcolsep}{3pt}
\renewcommand{\arraystretch}{1.0}
\begin{tabular}{@{}l>{\raggedright\arraybackslash}p{0.62\columnwidth}@{}}
\toprule
\textbf{Symbol} & \textbf{Meaning} \\
\midrule
$\pi_\theta,\pi_T,\pi_{\mathrm{ref}}$ & student / teacher / reference policy \\
$\pi_b$ & behaviour policy ($=\pi_\theta$ in on-policy GRPO) \\
$i,j \in \{1,\dots,G\}$ & sibling indices in a GRPO group of size $G$ \\
$t \in [0,T)$ & token position inside a trajectory of length $T$ \\
$\Adv^{(i)}$ & sequence-level GRPO advantage of trajectory $i$ \\
$\dt^{(i)}$ & teacher-student log-prob gap at $(i,t)$ \\
$\cti^{(i)}_t$ & estimated counterfactual token importance (P1) \\
$c_t \in (-1, 1)$ & signed credit (P1 output) \\
$\rhog,\rhoema$ & gate-active ratio and its EMA (P2) \\
$\lambda(t),\mu(t)$ & adaptive coefficients on CTI and KL (P2) \\
$\beta_+,\beta_-$ & sigmoid sharpness on positive / negative credit \\
$\tau$ & softmax temperature in the CTI estimator \\
$\theta_g$ & gate threshold for $\rhog$ \\
$\theta_{KL}$ & polarisation threshold (P3) \\
$\alpha$ & EMA decay of the P2 controller \\
$n_{\mathrm{warmup}}$ & controller warmup hold-out window \\
\bottomrule
\end{tabular}
\caption{Notation used throughout the paper.}
\label{tab:notation}
\end{table}

\end{document}